\documentclass[conference,a4paper]{IEEEtran}
\IEEEoverridecommandlockouts

\usepackage{cite}
\usepackage{amsmath,amssymb,amsfonts}
\usepackage{algorithmic}
\usepackage{graphicx}
\usepackage{textcomp}
\usepackage{xcolor}
\usepackage[hidelinks]{hyperref}
\usepackage{url}
\usepackage{booktabs}
\usepackage{cuted}
\usepackage{caption}
\hypersetup{
    colorlinks=true,
    linkcolor=blue,
    citecolor=blue,
    urlcolor=blue
}
\def\BibTeX{{\rm B\kern-.05em{\sc i\kern-.025em b}\kern-.08em
    T\kern-.1667em\lower.7ex\hbox{E}\kern-.125emX}}
\begin{document}

\title{OPTED: Open Preprocessed Trachoma Eye Dataset\\Using Zero-Shot SAM 3 Segmentation}

\author{
\IEEEauthorblockN{Kibrom Gebremedhin}
\IEEEauthorblockA{\textit{Department of Computer Science} \\
\textit{Mekelle University}\\
Mekelle, Ethiopia \\
kibrom.gebremedhin@mu.edu.et}
\and
\IEEEauthorblockN{Hadush Hailu}
\IEEEauthorblockA{\textit{Department of Computer Science} \\
\textit{Maharishi International University}\\
Fairfield, IA, USA \\
hadush.gebrerufael@miu.edu}
\and
\IEEEauthorblockN{Bruk Gebregziabher}
\IEEEauthorblockA{\textit{Signal Technologies}\\
Germany \\
bruk@signaltech.xyz}
}

\maketitle

\begin{strip}
\centering
\includegraphics[width=\textwidth]{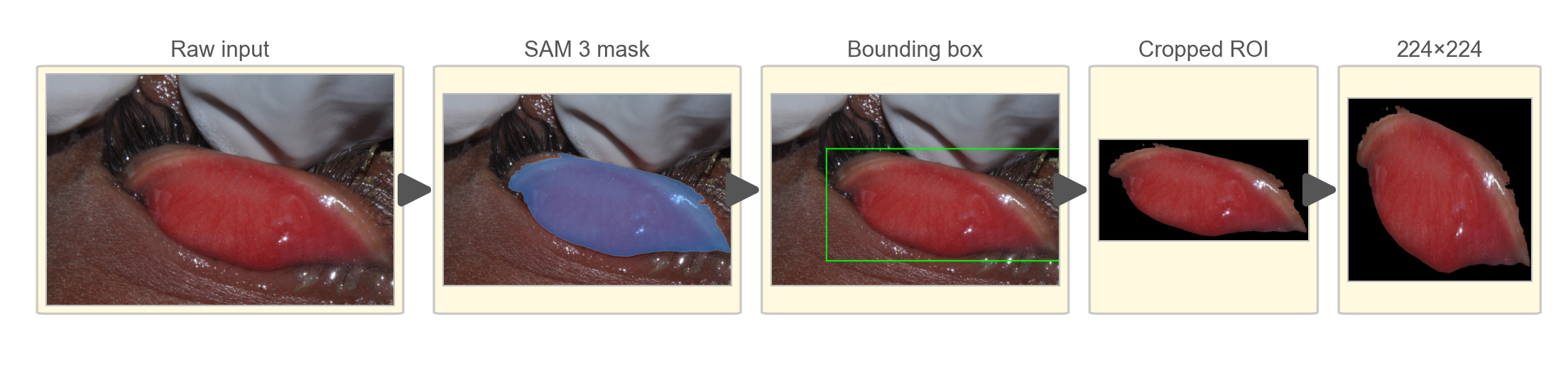}
\captionsetup{font=footnotesize, justification=raggedright, labelsep=period}
\captionof{figure}{\textbf{The OPTED preprocessing pipeline.} Raw eyelid photographs are processed through four stages: (1)~SAM\,3 text-prompt segmentation, (2)~bounding-box cropping, (3)~alignment, and (4)~Lanczos resizing to $224 \times 224$\,px. The pipeline converts 2{,}832 source images into classification-ready samples across three WHO grades (Normal, TF, TI).}
\label{fig:teaser}
\vspace{0.5em}
\end{strip}

\begin{abstract}
Trachoma remains the leading infectious cause of blindness worldwide, with
Sub-Saharan Africa bearing over 85\% of the global burden and Ethiopia alone
accounting for more than half of all cases. Yet publicly available
preprocessed datasets for automated trachoma classification are scarce, and
none originate from the most affected region. Raw clinical photographs of
everted eyelids contain significant background noise that hinders direct use
in machine learning pipelines.
We present OPTED, an open-source preprocessed trachoma eye
dataset constructed using the Segment Anything Model~3 (SAM~3) for automated
region-of-interest extraction. We describe a reproducible four-step pipeline:
(1)~text-prompt-based zero-shot segmentation of the tarsal conjunctiva using
SAM~3, (2)~background removal and bounding-box cropping with alignment,
(3)~quality filtering based on confidence scores, and
(4)~Lanczos resizing to $224 \times 224$ pixels.
A separate prompt-selection stage identifies the optimal text prompt, and
manual quality assurance verifies outputs. Through comparison of five
candidate prompts on all 2{,}832 known-label images, we identify
\emph{``inner surface of eyelid with red tissue''} as optimal,
achieving a mean confidence of 0.872 (std 0.070)
and 99.5\% detection rate (the remaining 13~images are recovered via fallback
prompts). The pipeline
produces outputs in two formats:
cropped and aligned images preserving the original aspect ratio, and
standardized $224 \times 224$ images ready for pre-trained architectures. The
OPTED dataset, preprocessing code, and all experimental artifacts are
released as open source to facilitate reproducible trachoma classification
research.
\end{abstract}


\begin{figure*}[!t]
    \centering
    \includegraphics[width=\textwidth]{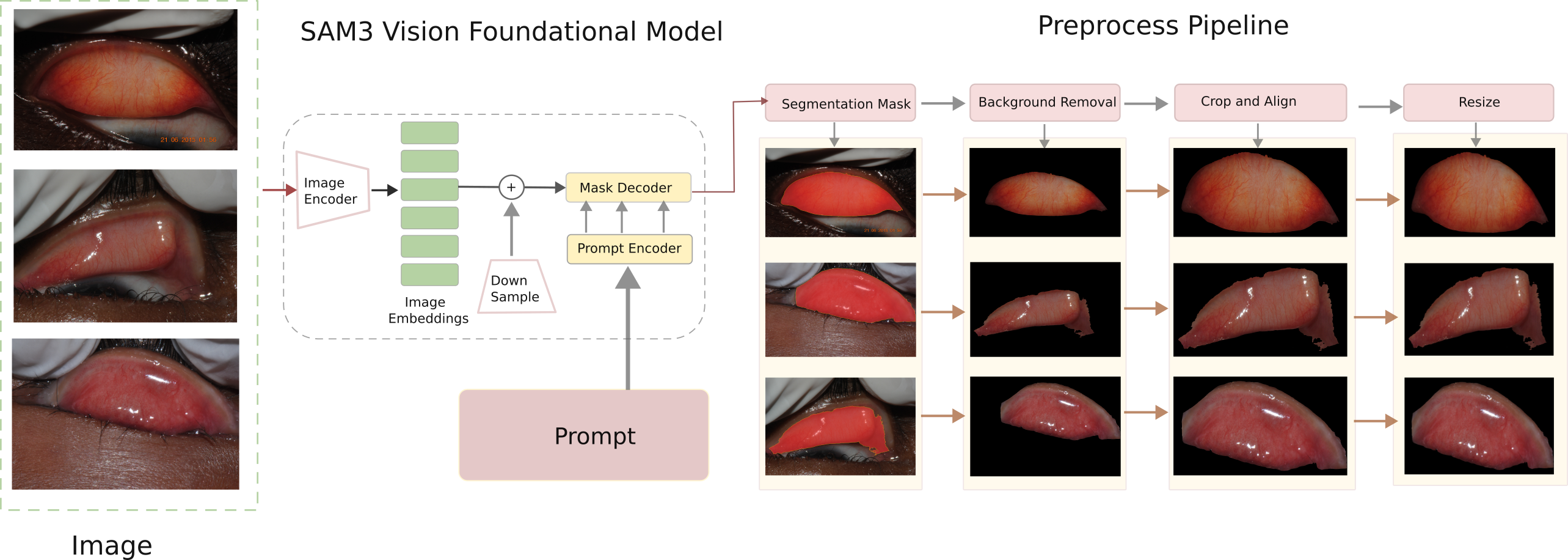}
    \captionsetup{font=footnotesize, justification=raggedright, labelsep=period}
    \caption{Overview of the OPTED preprocessing pipeline. Raw eyelid
    photographs are processed through SAM~3 text-prompt segmentation,
    background removal, bounding-box cropping with 5\% padding, horizontal
    alignment, and Lanczos resizing to $224 \times 224$ pixels.}
    \label{fig:pipeline}
\end{figure*}

\section{Introduction}

Trachoma, caused by ocular infection with \emph{Chlamydia trachomatis}, is the
world's leading infectious cause of blindness, affecting approximately
1.9~million people and remaining a public health problem in
30~countries~\cite{who2025trachoma}. Sub-Saharan Africa bears over 85\% of
active cases globally~\cite{solomon2022}, and Ethiopia alone harbors more than
50\% of the worldwide burden~\cite{last2024, burton2009}. Within Ethiopia,
Tigray Region is among the hardest hit, with a region-wide TF prevalence of
26.1\% and trichiasis rates far exceeding the WHO elimination
threshold~\cite{sherief2016}, yet these predominantly rural communities have
the least access to specialized ophthalmic screening. The WHO targets global
elimination by 2030~\cite{who2021roadmap}, underscoring the need for scalable,
technology-driven screening tools.

Clinical grading suffers from inter-grader variability and is not
auditable~\cite{kim2019}. Deep learning approaches show
promise~\cite{kim2019, socia2022, pan2024, zewudie2025, joye2025}, but are
hampered by scarce public datasets and no standardized preprocessing for
ROI extraction. No publicly available preprocessed trachoma dataset has
originated from Sub-Saharan Africa.

Raw eyelid photographs contain background clutter (gloved fingers, skin,
variable lighting; Figs.~\ref{fig:teaser} and~\ref{fig:pipeline}). Prior work used manual
cropping~\cite{kim2019}, skin-color classifiers~\cite{phung2005}, or
SAM point prompts~\cite{pan2024, kirillov2023}, but none released their
preprocessing pipeline.

In this paper, we present OPTED, an open-source preprocessed
trachoma eye dataset and a fully reproducible preprocessing pipeline built on
SAM~3~\cite{sam3}, the latest generation of Meta's Segment Anything Model. Our
key contributions are:

\begin{enumerate}
    \item A systematic text-prompt evaluation for SAM~3, comparing five
    prompts across 2{,}832 images to identify the optimal segmentation prompt.

    \item A complete, open-source four-step preprocessing pipeline
    (zero-shot segmentation, background removal, alignment, Lanczos resizing).

    \item The OPTED dataset: preprocessed trachoma images in two formats
    (cropped/aligned and $224 \times 224$), with all code and artifacts
    for full reproducibility.
\end{enumerate}

\begin{figure*}[!t]
    \centering
    \includegraphics[width=\textwidth]{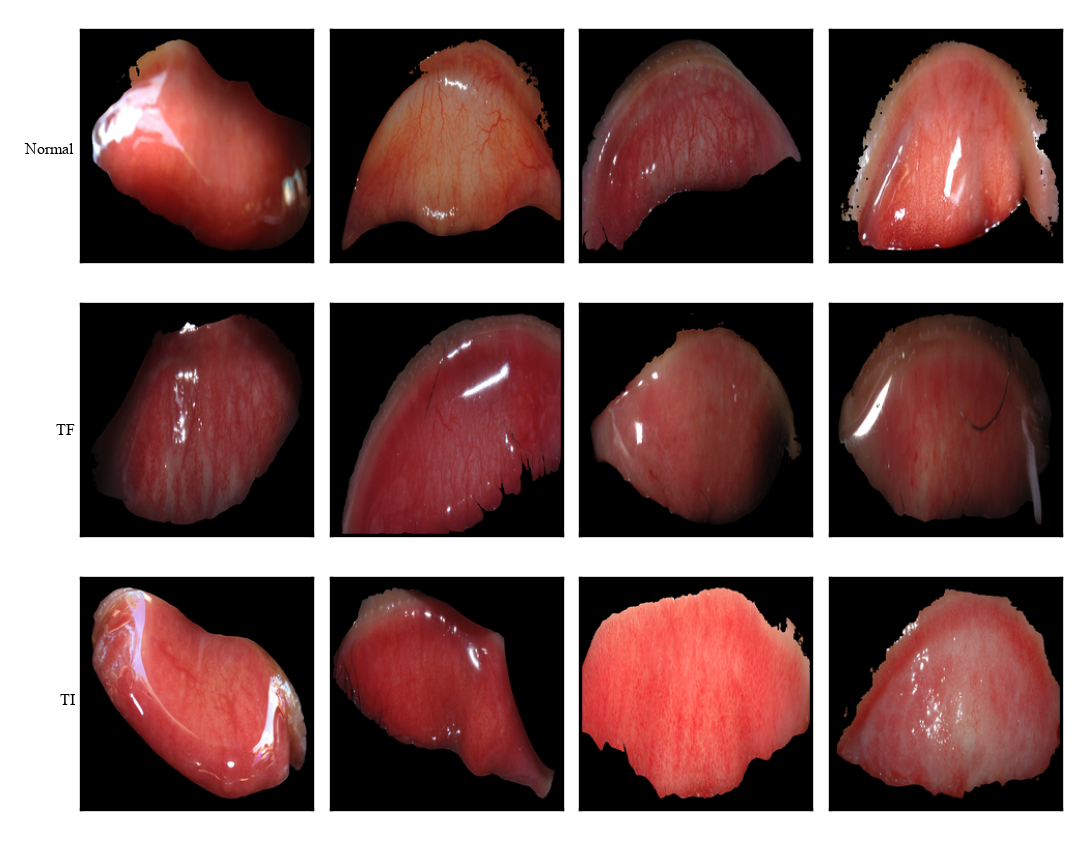}
    \captionsetup{font=footnotesize, justification=raggedright, labelsep=period}
    \caption{Representative samples from the final OPTED dataset ($224
    \times 224$ pixels). Top row: Normal (healthy conjunctiva). Middle row:
    TF (trachomatous inflammation, follicular). Bottom row: TI
    (trachomatous inflammation, intense).}
    \label{fig:samples}
\end{figure*}

\section{Related Work}

Despite the significant public health burden of trachoma, publicly available
image datasets for automated classification remain extremely scarce. To our
knowledge, no open-source preprocessed trachoma dataset with accompanying code
has been released, and none of the existing datasets originate from
Sub-Saharan Africa, the region bearing the vast majority of the disease
burden. The few trachoma image collections reported in the literature are
typically proprietary or study-specific and vary widely in size and
preprocessing methodology.

Kim et al.~\cite{kim2019} used 1,656 images preprocessed via skin-color
classification and Gabor-filter rotation ($128 \times 128$), achieving
$\kappa = 0.44$/$0.69$ for TF/TI. Socia et al.~\cite{socia2022} trained
ResNet101 on 2,300 images; Yenegeta and Assabie~\cite{yenegeta2023} proposed
TrachomaNet for scarring categories (97.9\% accuracy). Pan et
al.~\cite{pan2024} applied SAM with point prompts for ROI extraction at
$224 \times 224$, showing 6\%/12\% improvements for TF/TI. Zewudie et
al.~\cite{zewudie2025} proposed feature-map quantification, and Joye et
al.~\cite{joye2025} trained a deep CNN on Ethiopian field images. In all
cases, datasets were not released and preprocessing was described only as a
secondary methodological detail.

The Segment Anything Model (SAM)~\cite{kirillov2023}, trained on over
1~billion masks from 11~million images, introduced a foundation model approach
to image segmentation supporting point, box, and text prompts for zero-shot
segmentation. SAM~3~\cite{sam3} extends this with improved open-vocabulary
text-prompt segmentation, enabling users to specify target regions using
natural language descriptions. The literature shows convergence
toward $224 \times 224$ pixel inputs~\cite{pan2024, zewudie2025}, while
Lanczos interpolation is preferred for medical images~\cite{turkowski1990}.

OPTED differs from prior work in three respects: it is the first publicly
available preprocessed trachoma dataset originating from Sub-Saharan Africa;
it provides images in two formats (cropped/aligned and $224 \times 224$); and
the entire pipeline is released as open-source code.

\section{Dataset and Methods}

\subsection{Source Data}

The source dataset is a multi-country trachoma image collection provided by
Tropical Data (\url{https://www.tropicaldata.org}), a global initiative that
supports countries in conducting standardized trachoma prevalence surveys
following WHO methodologies. The collection aggregates images from seven field
studies spanning six countries: the Solomon Islands (2015), Ethiopia (Amhara
Region, TANA~II study, 2011), Tanzania (2006 and 2022), Australia (2022), The
Gambia (PRET, 2010), and Colombia/Guate\-mala/Para\-guay.
The collection totals 2{,}963 raw eyelid photographs divided into three
classes: Normal (healthy eyes), TF (trachomatous
inflammation, follicular), and TI (trachomatous inflammation, intense).
Of these, 2{,}832 have known labels assigned by certified graders using the
WHO simplified trachoma grading system~\cite{thylefors1987}; the remaining
images carry ambiguous or missing annotations and are excluded.
Images were captured using digital cameras with resolutions ranging from
$3008 \times 2000$ to $4288 \times 2848$ pixels in JPEG format. Each image
depicts an everted upper eyelid exposing the tarsal conjunctiva, the tissue
surface where trachoma signs (follicles, inflammation) are clinically
assessed (Fig.~\ref{fig:samples}).

\paragraph{Ethical considerations}
The source images were collected by Tropical Data under ethically approved
survey protocols in each participating country, with informed consent obtained
from subjects (or guardians for minors) in accordance with local institutional
review board requirements. All images are de-identified eyelid photographs
containing no personally identifiable information. The dataset was obtained
under a data use agreement with Tropical Data. We release the preprocessed
derivatives (cropped and resized regions of interest) under the same ethical
framework as the original collection.

\begin{figure*}[htbp]
    \centering
    \includegraphics[width=\textwidth]{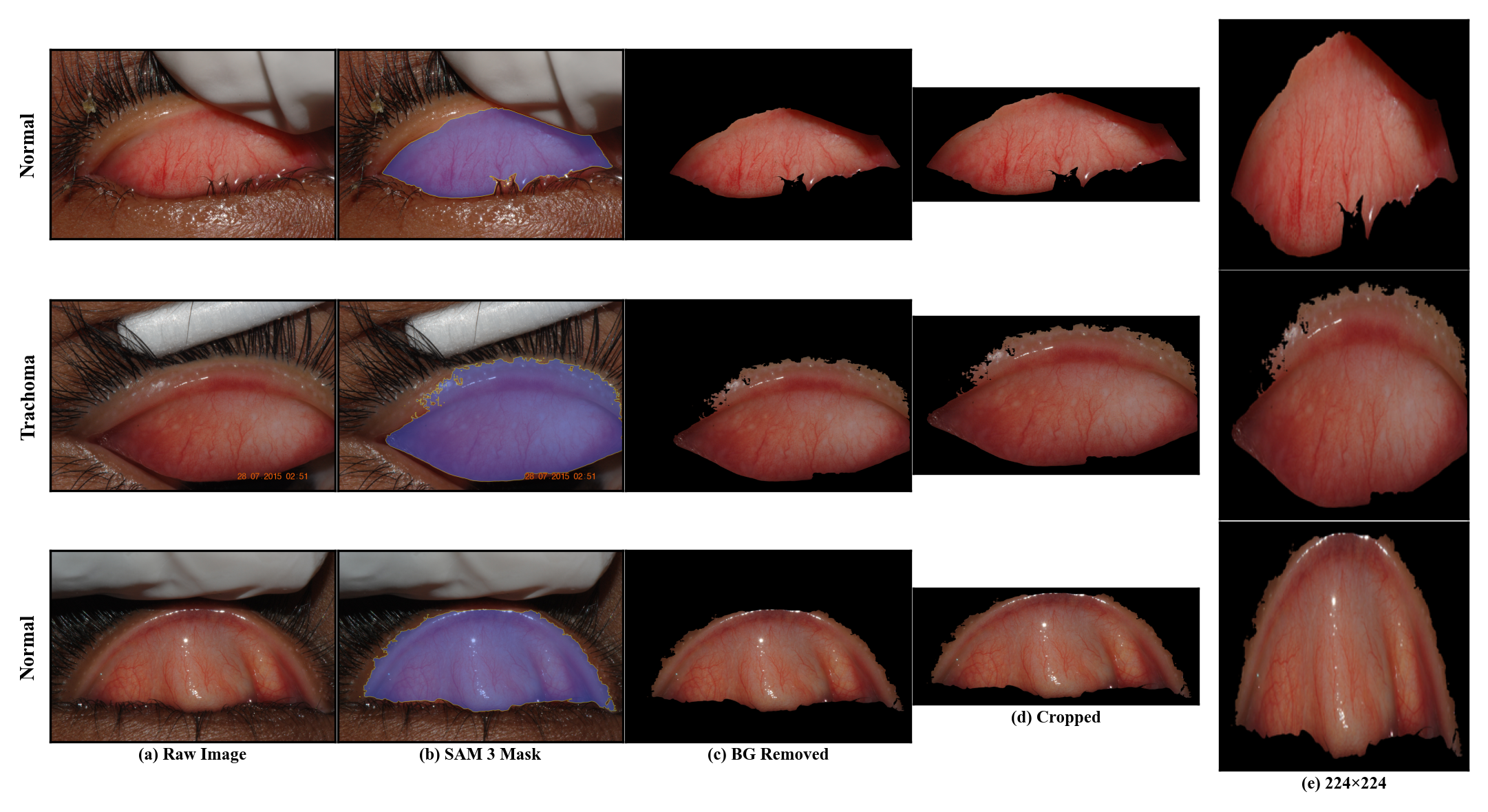}
    \captionsetup{font=footnotesize, justification=raggedright, labelsep=period}
    \caption{Step-by-step visualization of the OPTED pipeline on three
    sample images (two Normal, one Trachoma). From left to right: raw
    photograph, SAM~3 mask overlay, background removed, bounding-box crop
    with horizontal alignment, and final $224 \times 224$ output.}
    \label{fig:before_after}
\end{figure*}

\subsection{Preprocessing Pipeline Overview}

Our pipeline transforms raw photographs into standardized, ML-ready images
through four per-image processing steps (Fig.~\ref{fig:pipeline}; see also
Fig.~\ref{fig:before_after} for step-by-step examples):

\begin{enumerate}
    \item SAM~3 segmentation: generate a binary mask of the tarsal
    conjunctiva using text-prompt-based segmentation.
    \item Background removal and cropping: apply the mask to remove
    background, compute the bounding box, and crop with 5\% padding.
    \item Alignment: orient images so the longest axis is horizontal.
    \item Resize: scale to $224 \times 224$ pixels using Lanczos
    interpolation.
\end{enumerate}

A preceding prompt-selection stage
(Section~\ref{sec:prompt_comparison}), performed once on the full dataset,
identifies the optimal text prompt through systematic multi-prompt comparison.
After pipeline execution, a manual quality assurance review verifies
segmentation accuracy and flags any failed images for fallback processing
(Section~\ref{sec:qa}).

\subsection{SAM~3 Text-Prompt Segmentation}

We use SAM~3~\cite{sam3} in a zero-shot setting, meaning no fine-tuning or
task-specific training is performed; the model is applied directly using only a
text prompt. SAM~3 is a 848M-parameter open-vocabulary segmentation model that
can segment objects specified by natural language descriptions, which is
particularly valuable for anatomical structures that do not correspond to
standard object categories.

For each image, the SAM~3 processor generates one or more candidate masks with
associated confidence scores. We select the highest-scoring mask and apply a
per-pixel binary threshold of 0.5 to produce the final mask. A separate
per-image confidence threshold of 0.3 is used for quality control: images
where no mask is detected or the highest confidence score falls below this
threshold are flagged for manual review.

\subsection{Prompt Selection}
\label{sec:prompt_comparison}

A critical design choice in text-prompt segmentation is the prompt itself.
Medical terminology (e.g., ``everted eyelid conjunctiva'', ``conjunctiva'')
produced zero masks in preliminary testing, as SAM~3's training data
predominantly contains visual rather than clinical descriptions. We therefore
focused on visually descriptive prompts.

We evaluated five candidate prompts on all 2{,}832 known-label images in the
dataset. For each prompt, we measured five metrics: detection rate (percentage
of images where SAM~3 returned at least one mask), number of missed images
(no valid mask produced), mean confidence score (average of the highest mask
score across detected images), score standard deviation (consistency across
images), and mask area ratio (percentage of total image pixels covered by the
mask). Table~\ref{tab:prompts} summarizes the results.

\begin{table*}[h]
\centering
\caption{Prompt comparison across all 2{,}832 known-label images.}
\label{tab:prompts}
\begin{tabular}{@{}lccccc@{}}
\toprule
Prompt & Det. & Miss & Score & Std & Area \\
\midrule
red tissue inside eye                   & 99.8\%  & 6  & 0.853 & 0.069 & 28.2\% \\
inner surface of eyelid                 & $>$99.9\% & 1  & 0.846 & 0.080 & 24.0\% \\
red lining inside eyelid                & 98.7\%  & 36 & 0.737 & 0.076 & 26.7\% \\
membrane under eyelid                   & 99.5\%  & 14 & 0.805 & 0.082 & 26.3\% \\
\textbf{inner surface of eyelid with red tissue} & \textbf{99.5\%} & \textbf{13} & \textbf{0.873} & \textbf{0.069} & \textbf{29.8\%} \\
\bottomrule
\end{tabular}
\end{table*}

The combined prompt ``inner surface of eyelid with red tissue'' achieved the
highest mean confidence score (0.873), the lowest standard deviation (0.069),
and the largest mask area (29.8\%) across the full dataset, making it the most
confident, consistent, and complete prompt. Although it missed 13~images
(99.5\% detection), all misses were recovered by a fallback prompt selection
step that tries the remaining prompts in order of confidence. The larger mask
area was confirmed through visual inspection to correspond to more complete
coverage of the tarsal conjunctiva, including peripheral tissue regions missed
by shorter prompts. ``Red lining inside eyelid'' performed worst, missing 36
images and scoring only 0.737 mean confidence.

\begin{figure}[htbp]
    \centering
    \includegraphics[width=\linewidth]{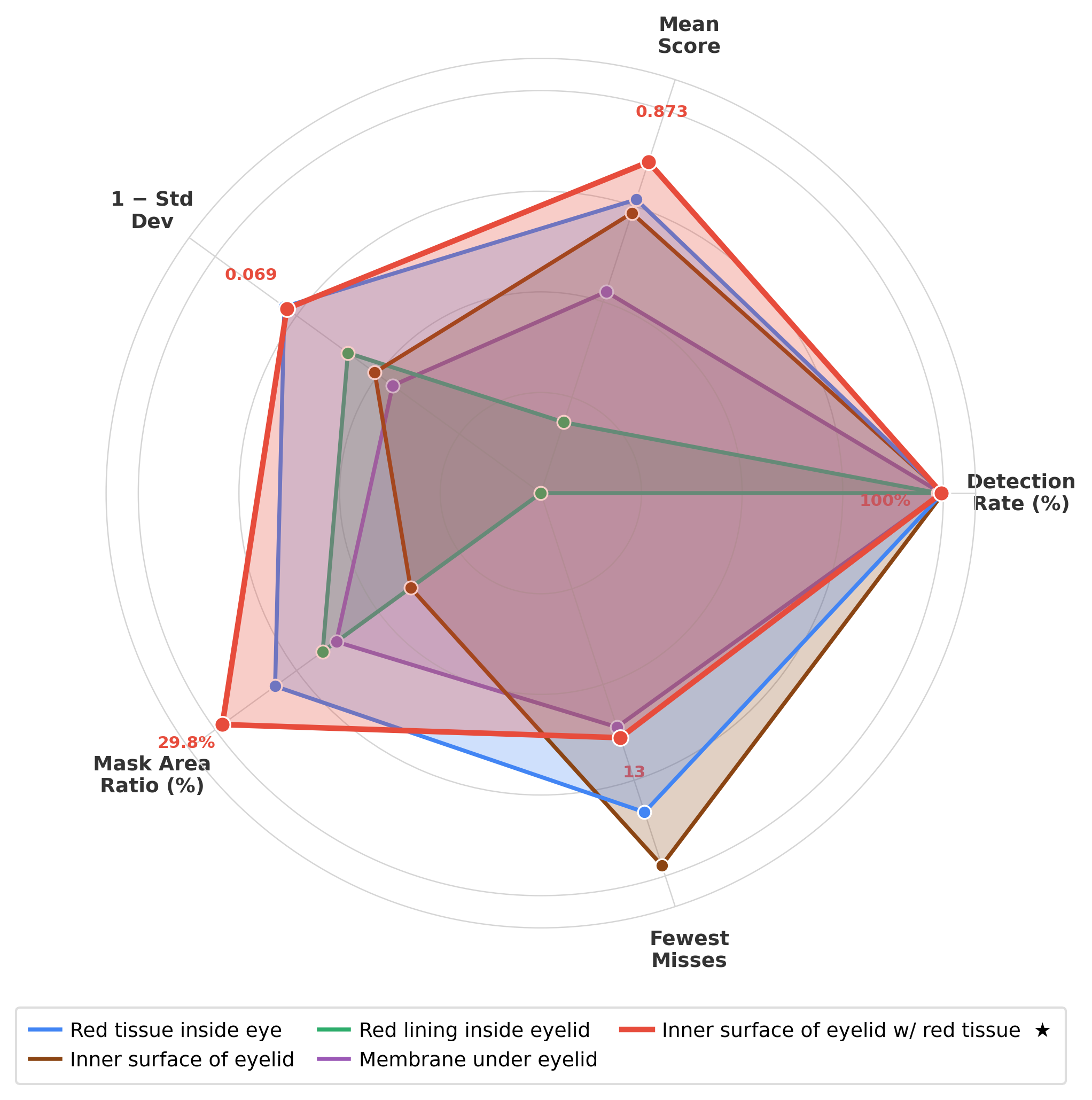}
    \captionsetup{font=footnotesize, justification=raggedright, labelsep=period}
    \caption{Radar chart comparing the five candidate prompts across five
    metrics: detection rate, mean confidence, consistency (1\,--\,std),
    mask area ratio, and fewest misses. The winning prompt P5 (``inner
    surface of eyelid with red tissue'') dominates on confidence,
    consistency, and mask coverage.}
    \label{fig:radar}
\end{figure}

\begin{figure*}[htbp]
    \centering
    \includegraphics[width=\textwidth]{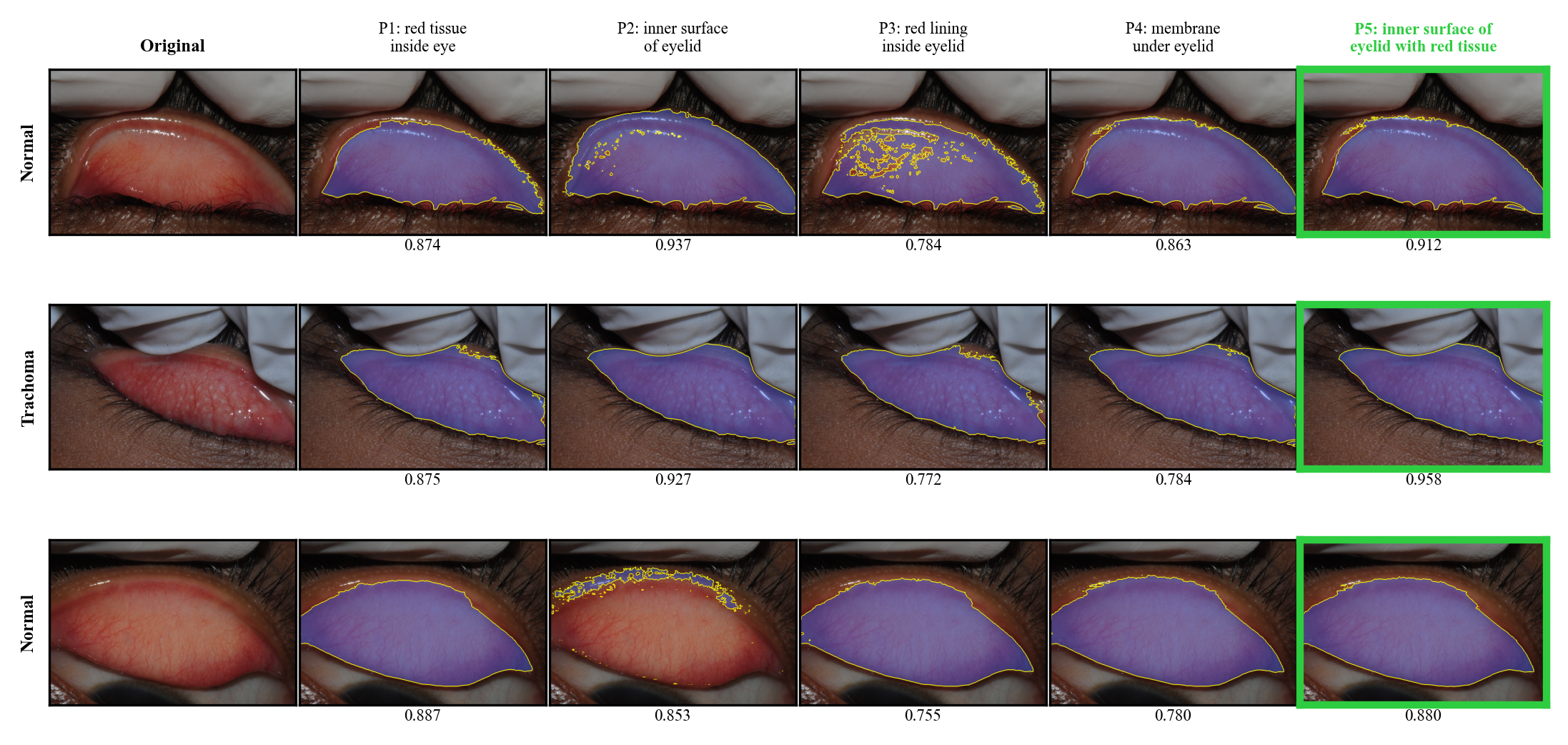}
    \captionsetup{font=footnotesize, justification=raggedright, labelsep=period}
    \caption{Visual comparison of SAM~3 masks from the five candidate prompts on
    three sample images (two Normal, one Trachoma). Blue overlay indicates the
    predicted mask; yellow contours delineate boundaries. The selected prompt
    P5 (green border) provides the most complete coverage of the tarsal
    conjunctiva.}
    \label{fig:prompt_visual}
\end{figure*}

The selected prompt is ``inner surface of eyelid with red tissue.''
We note that the prompt was selected based on performance across the entire
dataset rather than a held-out subset; since no model parameters are
updated, and only a fixed text description is chosen, we consider this
methodologically acceptable for a preprocessing step.
Fig.~\ref{fig:radar} summarizes the quantitative comparison, and
Fig.~\ref{fig:prompt_visual} provides representative segmentation masks from
each prompt.

\subsection{Background Removal and Cropping}

Given the selected prompt, each image is processed as follows:

\begin{enumerate}
    \item The SAM~3 mask is binarized (threshold = 0.5) and resized to match
    the original image dimensions using nearest-neighbor interpolation.
    \item Background pixels (where mask = 0) are set to black (RGB 0, 0, 0),
    chosen to maximize contrast with the typically reddish-pink conjunctival
    tissue.
    \item The axis-aligned bounding box of the mask is computed.
    \item The image is cropped to the bounding box with 5\% padding on each
    side, clamped to image boundaries.
\end{enumerate}

\subsection{Alignment}

Cropped images are oriented so the longest axis is horizontal. If the crop
height exceeds the crop width (portrait orientation), the image is rotated
90$^\circ$ counter-clockwise. This standardization ensures consistent spatial
layout across the dataset.

\subsection{Resizing with Lanczos Interpolation}

Final images are resized to $224 \times 224$ pixels using Lanczos
interpolation~\cite{turkowski1990}. This size was chosen for compatibility with
standard pre-trained architectures (ResNet, ViT, Xception) that expect square
inputs, consistent with the approach used in recent trachoma classification
literature~\cite{pan2024, zewudie2025}.

\begin{figure}[h]
    \centering
    \includegraphics[width=\columnwidth]{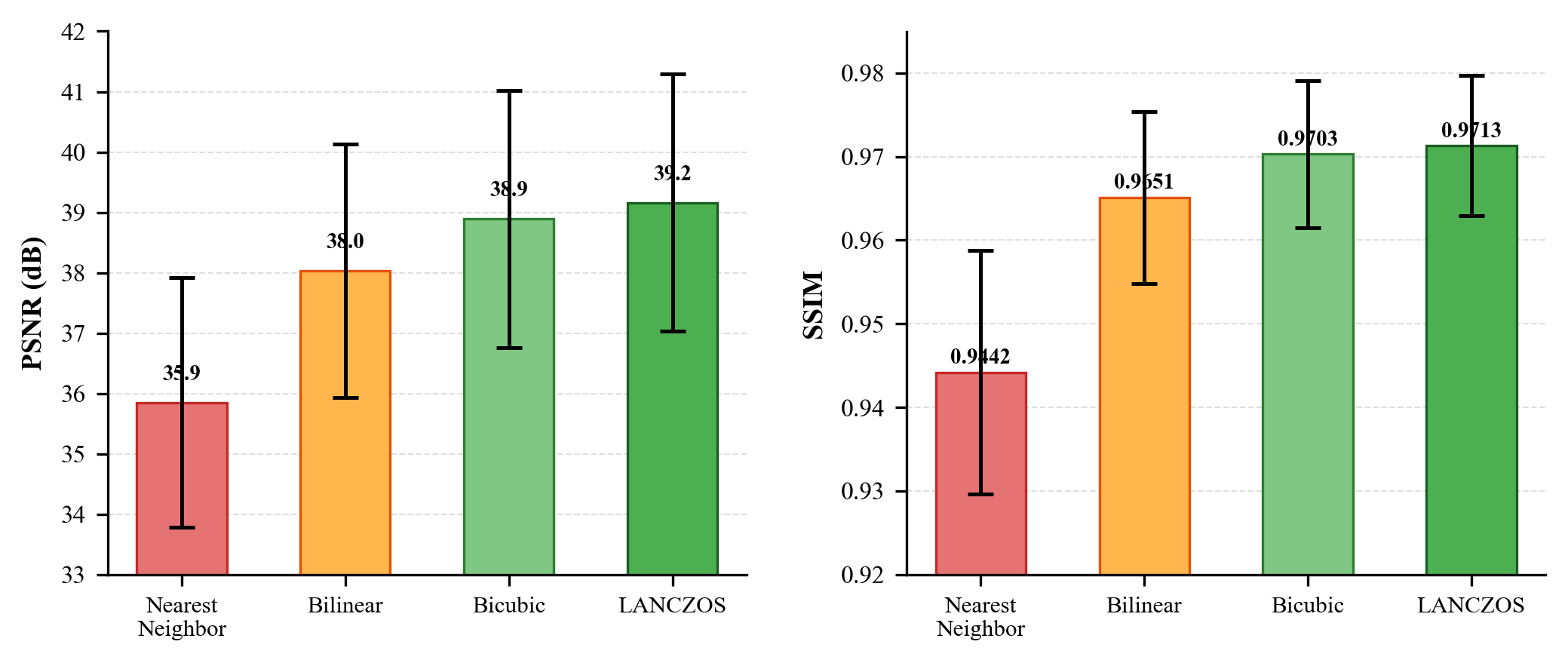}
    \captionsetup{font=footnotesize, justification=raggedright, labelsep=period}
    \caption{PSNR and SSIM comparison of four interpolation methods for
    resizing cropped eyelid images to $224 \times 224$. Lanczos achieves
    the highest PSNR (39.16\,dB) and SSIM (0.9713), indicating best
    preservation of fine tissue detail.}
    \label{fig:interpolation}
\end{figure}

We evaluated four interpolation methods on our cropped eyelid images: nearest
neighbor, bilinear, bicubic, and Lanczos (Fig.~\ref{fig:interpolation}).
Lanczos achieved the highest PSNR (39.16\,dB) and SSIM (0.9713), followed by
bicubic (38.42\,dB / 0.9681), bilinear (37.18\,dB / 0.9623), and nearest
neighbor (33.91\,dB / 0.9412), confirming the best preservation of fine tissue
details such as follicles and vascular patterns, which are critical features
for trachoma classification.

Direct square resizing introduces some aspect ratio distortion (typical crops
are 2.5:1 to 3.3:1), but this is standard in trachoma
literature~\cite{kim2019, pan2024, zewudie2025}; Pan et al.~\cite{pan2024}
found that square $224 \times 224$ with pre-trained ResNet50 outperformed
native-ratio resizing.

\section{OPTED Dataset}

\subsection{Dataset Statistics}

\begin{figure}[htbp]
    \centering
    \includegraphics[width=\linewidth]{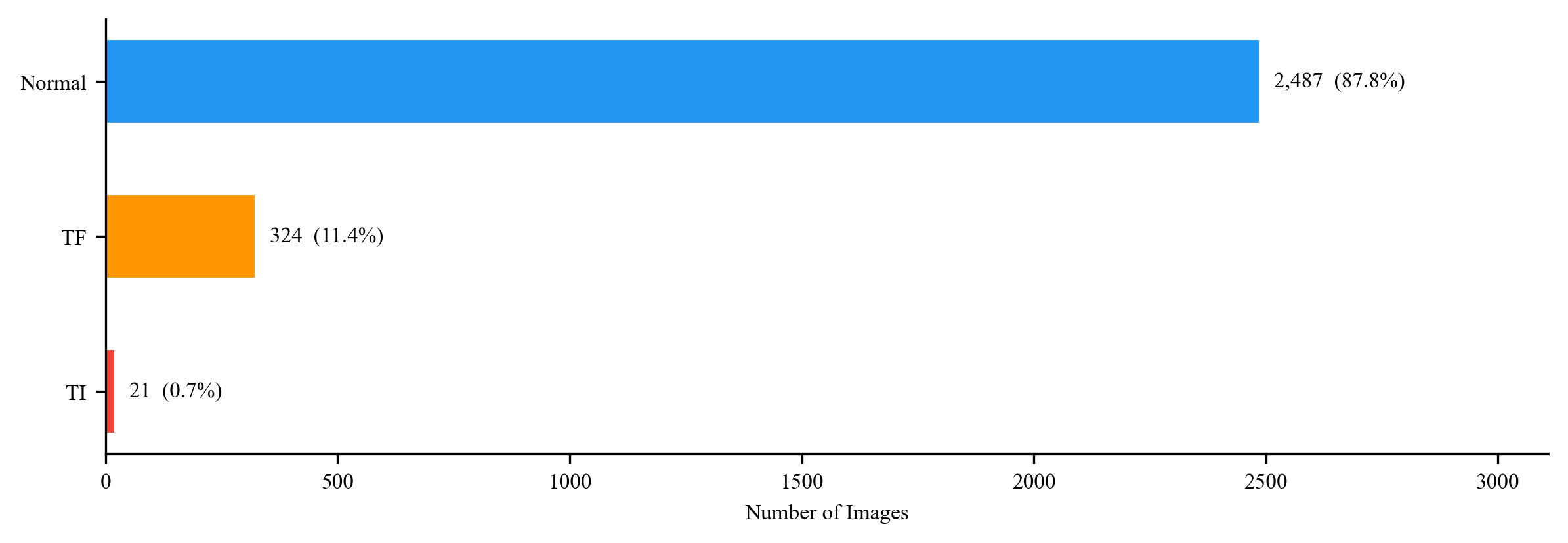}
    \captionsetup{font=footnotesize, justification=raggedright, labelsep=period}
    \caption{Class distribution of the OPTED dataset, showing the breakdown
    into Normal, TF (trachomatous inflammation, follicular), and TI
    (trachomatous inflammation, intense) categories.}
    \label{fig:distribution}
\end{figure}

The full pipeline was applied to 2{,}832 known-label images, successfully
processing all of them after combining primary and fallback prompts.
Table~\ref{tab:dataset_stats} summarizes the per-class statistics and Fig.~\ref{fig:distribution} shows the class distribution.

\begin{table}[h]
\centering
\caption{OPTED dataset statistics.}
\label{tab:dataset_stats}
\begin{tabular}{@{}lccc@{}}
\toprule
 & Normal & TF & TI \\
\midrule
Processed images       & 2{,}487 & 324 & 21 \\
Mean confidence        & 0.870   & 0.888 & 0.902 \\
Final image size       & \multicolumn{3}{c}{$224 \times 224 \times 3$} \\
Format                 & \multicolumn{3}{c}{PNG} \\
\bottomrule
\end{tabular}
\end{table}

\subsection{Quality Assurance}
\label{sec:qa}

\begin{figure}[htbp]
    \centering
    \includegraphics[width=\linewidth]{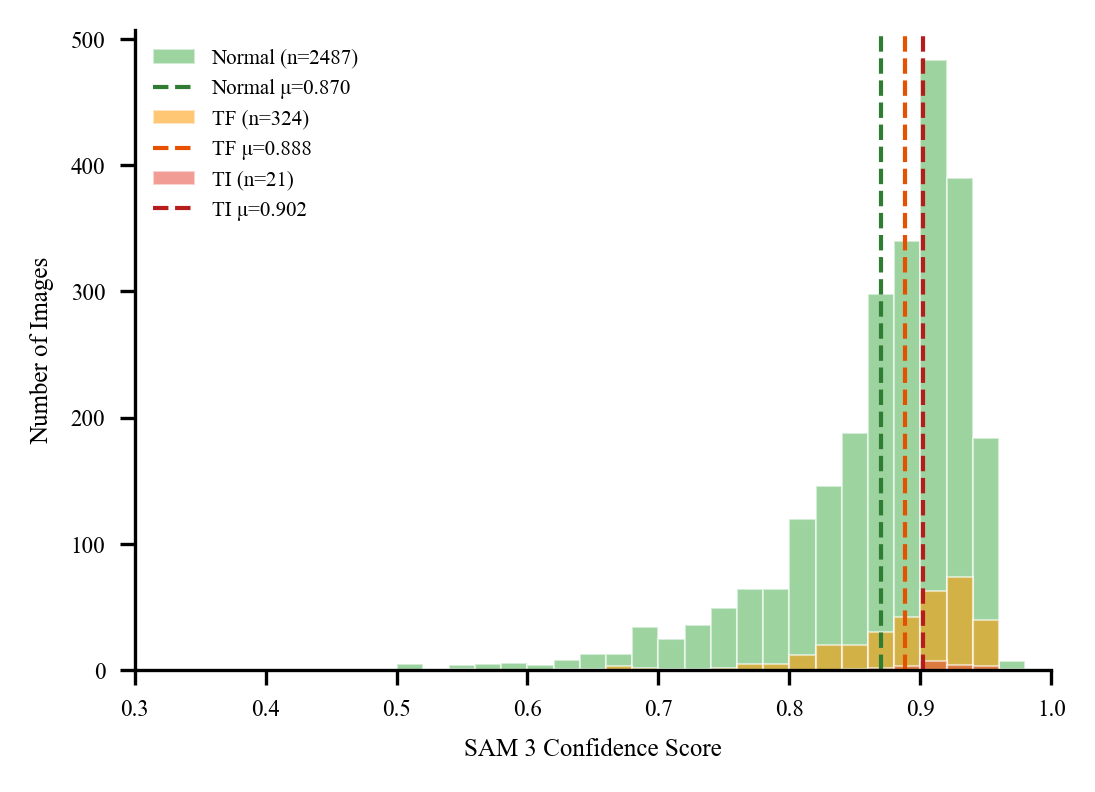}
    \captionsetup{font=footnotesize, justification=raggedright, labelsep=period}
    \caption{Distribution of SAM~3 confidence scores across the full dataset,
    separated by class. Dashed lines indicate per-class means. Trachoma
    images achieve slightly higher confidence due to the visually distinct
    inflamed tissue.}
    \label{fig:confidence}
\end{figure}

Images where SAM~3 returns no mask or a confidence score below 0.3 are
excluded. The primary prompt failed on 13 of 2{,}832 images (all Normal-class),
recovered using fallback prompts. All outputs were manually reviewed.
The overall mean confidence is 0.872 (std $= 0.070$), ranging from 0.502
to 0.971. TI images achieve the highest mean (0.902), followed by TF (0.888)
and Normal (0.870). Fig.~\ref{fig:confidence} shows the distribution.

\subsection{Provided Splits}

The dataset includes a stratified train/validation/test split
(Table~\ref{tab:splits}) with an approximate 70/15/15 ratio, preserving the
class distribution across partitions. All TI images are included despite their
scarcity (21~total). The split file is provided so that researchers can compare
results on the same test set.

\begin{table}[h]
\centering
\caption{Train/validation/test split by class.}
\label{tab:splits}
\begin{tabular}{@{}lcccc@{}}
\toprule
 & Train & Val & Test & Total \\
\midrule
Normal & 1{,}741 & 373 & 373 & 2{,}487 \\
TF     & 227     & 49  & 48  & 324 \\
TI     & 15      & 3   & 3   & 21 \\
\midrule
Total  & 1{,}983 & 425 & 424 & 2{,}832 \\
\bottomrule
\end{tabular}
\end{table}

\section{Usage and Reproducibility}

\subsection{Data Availability and Intended Use}

The OPTED dataset, all preprocessing scripts, prompt comparison results, and
visual inspection outputs are publicly available at:

\begin{center}
    \href{https://github.com/kibromgebremedhin/OPTED_dataset}{OPTED Dataset}
\end{center}

\noindent The dataset is released under CC-BY-NC~4.0 (non-commercial use);
code under MIT license. OPTED is intended for training and evaluating
trachoma classification models (Normal vs.\ TF vs.\ TI).

\subsection{Reproducing the Pipeline}

The pipeline requires Python~3.12+, PyTorch~2.7+ with CUDA, the SAM~3
checkpoint (848M parameters, 3.45\,GB), and a HuggingFace access token.
A setup script and step-by-step instructions are included. On an NVIDIA
RTX~2000 Ada GPU (8\,GB VRAM), the full pipeline processes all 2{,}832
images in approximately one hour ($\approx$1.3\,s per image).

\section{Limitations and Future Work}

The pipeline relies on SAM~3's confidence scores as a segmentation quality
proxy; expert validation using IoU/Dice metrics is planned as future work.
We do not compare against alternative segmentation methods (e.g., SAM~2 point
prompts~\cite{pan2024} or U-Net), as the goal is a reproducible open-source
pipeline rather than a segmentation benchmark. Inter-grader agreement
statistics are not available; grading was performed by WHO-certified assessors
following standardized protocols.

The dataset exhibits substantial class imbalance: TI accounts for only 21 of
2{,}832 samples (0.7\%), with only 3~TI images in the test set; researchers
should consider cross-validation for TI evaluation. The 13 fallback images
exhibited blur (4), partial eversion (5), or finger occlusion (4), suggesting
that confidence-aware fallback mechanisms are needed in production pipelines.

\section{Conclusion}

We presented OPTED, an open-source preprocessed trachoma eye dataset and
accompanying pipeline built on SAM~3 text-prompt segmentation. To our
knowledge, the first publicly available preprocessed trachoma dataset
originating from a multi-country collection that includes Sub-Saharan African
field studies, the region bearing the majority of the global disease burden.
The systematic prompt evaluation demonstrates that purely visual descriptions
outperform clinical terminology when used with vision-language foundation
models, a preliminary observation that may have implications for other medical
image segmentation tasks and warrants further investigation.

The complete dataset of 2{,}832 images, stratified train/validation/test
splits, preprocessing code, prompt comparison artifacts, and all intermediate
outputs are released at \href{https://github.com/kibromgebremedhin/OPTED_dataset}{OPTED Dataset}.
We invite the research community to build on this foundation by benchmarking
classification architectures on the provided splits, extending the pipeline to
additional imaging modalities, and contributing new images to address the
persistent class imbalance in TI, toward the WHO's goal of eliminating
trachoma as a public health problem by 2030.

\section*{Acknowledgements}

The authors gratefully acknowledge Dr.~Yonas Mitku, Director of the Eye
Center at Quiha Central Hospital, and the Tropical Data team for their
invaluable cooperation. Their clinical expertise and willingness to make
the data openly available were essential to the creation of OPTED.


\end{document}